\documentclass[10pt,twocolumn,letterpaper]{article}

\usepackage[pagenumbers]{cvpr}      %

\usepackage{multirow}
\usepackage{siunitx}  %
\usepackage{algorithm,algpseudocode}  %
\usepackage{tikz}     %
\usetikzlibrary{arrows.meta,calc,positioning,decorations.pathreplacing,backgrounds}

\newcommand{\topfive}[1]{\,{\scriptsize(#1)}}  %

\definecolor{cvprblue}{rgb}{0.21,0.49,0.74}
\usepackage[pagebackref,breaklinks,colorlinks,allcolors=cvprblue]{hyperref}

\def\confName{CVPR}
\def\confYear{2027}

\title{PixLUT: Nanosecond ImageNet Classification with LUT-based Convolutional Architectures on Pure FPGA Fabric}

\author{Hans Farrell Soegeng\\
Nanyang Technological University\\
{\tt\small hans0048@e.ntu.edu.sg}
\and
Sarthak Ketanbhai Modi\\
Nanyang Technological University\\
{\tt\small sarthak010@e.ntu.edu.sg}
\and
Thomas Peyrin \\
Nanyang Technological University\\
{\tt\small thomas.peyrin@ntu.edu.sg}
}

\begin{document}
\maketitle

\begin{abstract}
\label{sec:abstract}
Low-latency, high-throughput vision inference has become increasingly important for real-time applications.
FPGAs target this regime. When every neuron is mapped onto the lookup tables (LUTs) that form the FPGA's native logic, a fully unrolled network runs with massive parallelism, nanosecond latency and low power.
However, such fully unrolled LUT-based networks have been limited to flat architectures (multilayer perceptrons and tree ensembles over a flattened input) without the spatial understanding that convolutions provide.
In this work, we present PixLUT, a convolutional network that is fully unrolled on pure FPGA fabric, with no DSPs, block RAM or multipliers, and trained end to end with differentiable LUT neurons whose inputs are selected through structured sparsity.
Averaged over three seeds, PixLUT achieves 74.9\% top-1 accuracy on CIFAR-10, 48.6\% on CIFAR-100, 27.2\% on Tiny ImageNet and 11.1\% on ImageNet-1k ($32\times32$), classifying one image per clock at 200--204\,MHz with 59--75\,ns latency.
To the best of our knowledge, this is the first convolutional network built entirely from native FPGA logic and fully unrolled at one image per clock to scale up to ImageNet-1k, routed on a single FPGA.

\end{abstract}

\section{Introduction}
\label{sec:intro}

A decade of work on efficient architectures~\cite{howard2017mobilenets,zhang2018shufflenet}, quantization~\cite{rastegari2016xnor,esser2020lsq} and pruning~\cite{han2015learning,han2016deepcompression} has made visual recognition cheap.
Yet a class of vision systems needs it fast in a stricter sense, with a decision for every incoming sample, at the rate of the sensor, within nanoseconds.
Particle-physics triggers must filter collisions arriving at 40\,MHz~\cite{duarte2018hls4ml}, and high-speed inspection and closed-loop control face similar budgets.
Efficient networks still execute as sequences of multiply-accumulate operations on processors built for batched throughput, which puts this regime out of reach.

FPGAs can meet this budget, but only if the entire network can be implemented in the FPGA's native logic, lookup tables (LUTs).
Most FPGA accelerators still emulate multiply-accumulate arithmetic with on-chip memory, time-multiplexed over the input, and stay at microsecond latency~\cite{umuroglu2017finn,ghasemzadeh2018rebnet}.
LUT networks remove most of the arithmetic. The network is trained into small truth tables mapped onto the FPGA's native lookup tables (LUTs), fully unrolled, and a new input enters every clock cycle~\cite{umuroglu2020logicnets,andronic2024neuralut,bacellar2024dwn,khataei2025treelut,hoang2026kanele}.

These networks are flat, using multilayer perceptrons, tree ensembles or logic-gate layers~\cite{petersen2022difflogic,fojcik2025lilogic,buhrer2026bitlogic} over a flattened input, unaware of spatial structure and channels, and they degrade beyond image datasets with only 1 input channel such as MNIST.
Convolutional logic-gate networks, such as~\cite{petersen2024convdlgn,wang2026compactlogic}, learn two-input Boolean gates rather than the FPGA's native 6-input unit, so their cost is reported in gates, not in the LUTs of a routed circuit.
A streaming LUT CNN method~\cite{liang2026fpgn} does use 6-input LUTs, but reuses its hardware across positions and so gives up the one image per clock property.

Designing a LUT-based convolutional network faces 3 main problems.
First, a LUT reads six bits, but a $3\times3$ window over 64 channels of 2-bit activations offers 1{,}152, so each neuron must learn which few to read, across channels and positions (\emph{fan-in}).
Second, each LUT computes an arbitrary Boolean function of quantized inputs, which gradients reach only through approximation (\emph{discrete learning}).
Third, unrolling gives every position its own hardware, so a vision network becomes hundreds of thousands of LUTs that must route and close timing across a multi-die device, exactly where the learned all-to-all wiring of flat networks fails (\emph{scale}).

We present PixLUT, a convolutional network in which every neuron reads six bits and is one 6-input LUT per output bit.
Each output channel learns a sparse, cross-channel receptive field by annealed top-$k$ selection, extending spatial top-$k$ pruning~\cite{modi2026topk} from positions within a kernel to the full receptive field, and a nonlinear function of it by a small sub-network that is tabulated after training~\cite{andronic2024neuralut}.
The truth table and its inputs are shared across positions, a weightless LUT head classifies~\cite{bacellar2024dwn}, and the model compiles bit-exactly into a fully unrolled, pipelined netlist.
Locality-preserving layout (block-local head wiring, per-die reduction) lets networks of up to 696k LUTs route on one FPGA.

On an AMD Alveo U250, with seven LUT and logic-gate networks re-implemented under the same flow, PixLUT reaches 74.9\%, 48.6\%, 27.2\% and 11.1\% top-1 on CIFAR-10, CIFAR-100, Tiny ImageNet and ImageNet at $32\times32$ (mean of three seeds), at 59--75\,ns latency and one image per clock.
This is 1.2--3.7$\times$ the top-1 accuracy of the best routed baseline, a margin that widens with task difficulty, while the baselines' larger models fail to train or route.

Our contributions are:
\begin{itemize}
  \item A convolutional LUT backbone in which each output channel learns which few taps to read, across channels as well as positions, by annealed top-$k$ selection shared across positions; following NeuraLUT~\cite{andronic2024neuralut}, each node's function is a small sub-network over its selected taps, initialised to the masked convolution and tabulated into one LUT per output bit (\cref{sec:backbone}).
  \item A bit-exact compilation of the whole network, through the argmax, into a fully unrolled, pipelined netlist of native FPGA logic only (LUTs, flip-flops and a few carry chains, with no DSPs or block RAM), verified against the trained model on the full evaluation split, with a locality-preserving layout (learned but block-local head wiring, per-die reduction) that routes designs of up to 696k LUTs on one multi-die FPGA at 200--204\,MHz (\cref{sec:compile,sec:scale}).
  \item A comparison that runs seven LUT and logic-gate networks through the same device and flow as PixLUT, each ending at the registered class index.
  PixLUT is the most accurate placed-and-routed design in this comparison on all six datasets, with 1.2--3.7$\times$ the top-1 accuracy of the best routed baseline from CIFAR-10 to ImageNet-1k, and is, to our knowledge, the first convolutional network implemented entirely in native FPGA logic, fully unrolled at one image per clock and routed on a single FPGA, to be evaluated on Tiny ImageNet and ImageNet-1k ($32\times32$) (\cref{sec:experiments}).
  \item Code and the routed designs, released upon acceptance; architectures and training configurations are in the appendix (\cref{sec:arch,sec:recipe}).
\end{itemize}

\section{Related work}
\label{sec:related}

We review FPGA inference that still relies on arithmetic, networks that replace arithmetic with the FPGA's native LUTs, and the structured sparsity our convolution builds on.

\paragraph{Arithmetic-based FPGA inference.}
Binary and low-bit networks~\cite{hubara2016bnn,rastegari2016xnor} reduce multiplications to XNOR and popcount, and FPGA frameworks map them onto dataflow pipelines that keep weights in block RAM and fold the computation over the input~\cite{umuroglu2017finn,blott2018finnr,ghasemzadeh2018rebnet}; hls4ml~\cite{duarte2018hls4ml} maps small networks onto DSP blocks for particle-physics triggers.
Tridgell \etal~\cite{tridgell2019unrolling} hard-code a ternary CNN into logic but stream one pixel per cycle, and LUTNet~\cite{wang2019lutnet,Wang2019LUTNetLF} and Logic Shrinkage~\cite{wang2022logicshrinkage} learn the XNORs into LUTs but keep the binarized datapath and its memory for most layers.
All keep arithmetic and memory in the datapath, which at vision scale means folding and microsecond latency (\cref{tab:regimes}).

\paragraph{Arithmetic-free LUT and logic networks.}
LogicNets~\cite{umuroglu2020logicnets} trains sparse, quantized neurons that are enumerated into truth tables and unrolled into LUTs, one sample per clock.
PolyLUT~\cite{andronic2023polylut,andronic2025polylut}, PolyLUT-Add~\cite{lou2024polylutadd}, NeuraLUT~\cite{andronic2024neuralut,andronic2025assemble} and KANEL\'E~\cite{hoang2026kanele} enrich each neuron's function, DWN~\cite{bacellar2024dwn} learns the tables and their wiring, and TreeLUT~\cite{khataei2025treelut} unrolls boosted trees.
Logic-gate networks learn a two-input gate per node~\cite{petersen2022difflogic,fojcik2025lilogic}.

Convolutional variants make each filter a small truth table for verification~\cite{benamira2024ttnet,benamira2023ttnet}, or build kernels from two-input gates over one or two channels~\cite{petersen2024convdlgn,wang2026compactlogic}.
Concurrently, FPGN~\cite{liang2026fpgn} builds a CNN from 6-input LUTs with structured wiring, but streams each image through reused hardware over many cycles, giving up one image per clock and, with it, throughput and latency.
Outside FPGAs, vision uses lookup tables as caches in memory, read by a processor~\cite{jo2021srlut,tang2023lutnn}. PixLUT instead implements the vision logic with lookup tables directly.

\paragraph{Sparsity.}
Pruning keeps weights by magnitude, often gradually~\cite{zhu2017prune}, or in structured patterns such as N:M~\cite{zhou2021nm} or the top-$k$ positions of each kernel~\cite{modi2026topk}, to save operations on hardware that still computes. LUT networks instead need sparsity to bound each neuron's input so that it can be enumerated, with inputs fixed at random~\cite{umuroglu2020logicnets} or learned per node~\cite{wang2022logicshrinkage,bacellar2024dwn}.
PixLUT learns two to six of up to 720 inputs per neuron, across channels as well as positions.

\begin{table}[t]
\centering
\footnotesize
\caption{CIFAR-10 designs in other execution regimes, with each paper's own numbers and device (not re-implemented).
Throughput is images per second, and ``--'' marks values not reported.
$^a$Rebuilt by Logic Shrinkage~\cite{wang2022logicshrinkage}, post-implementation.
$^b$12{,}109\,GOp/s divided by the paper's 118.9\,MOp per image.
$^c$System latency, including PCIe; LUTs of the CNN alone.}
\label{tab:regimes}
\setlength{\tabcolsep}{1.8pt}
\begin{tabular}{lllrrrr}
\toprule
Method & Regime & Device & Acc.\,\% & LUTs & Latency & Img/s \\
\midrule
FINN~\cite{umuroglu2017finn}               & folded        & Z7045  & 80.1  & 46.3k  & 283\,\textmu s & 21.9k \\
FINN-R~\cite{blott2018finnr}               & folded        & VU9P   & 80.1  & 332.6k & --        & 102k$^b$ \\
ReBNet~\cite{ghasemzadeh2018rebnet}        & folded        & Z7020  & 87.0  & --     & --        & 2k \\
LUTNet$^a$~\cite{Wang2019LUTNetLF}        & partial       & KU115  & 84.6  & 291.3k & --    & -- \\
Logic Shrink.~\cite{wang2022logicshrinkage} & partial    & KU115  & 84.7  & 188.8k & --        & -- \\
Tridgell~\cite{tridgell2019unrolling} & per-pixel & VU9P   & 90.9  & 549.4k & 29\,\textmu s$^c$ & 122k \\
FPGN~\cite{liang2026fpgn}              & stream    & VP1902 & 82.9  & 2.46M  & 658\,ns   & 3.2M \\
\midrule
PixLUT                                      & unrolled      & U250   & 74.9  & 264.0k & 58.8\,ns  & 204M \\
\bottomrule
\end{tabular}
\end{table}

\section{Method}
\label{sec:method}

\paragraph{Preliminaries.}
An FPGA's logic fabric consists of 6-input lookup tables (LUTs), each storing 64 bits and realizing any Boolean function of its six inputs in a single level of logic.
A neuron that reads $k$ activations of $b$ bits is a function of $kb$ bits, whatever it computes internally (a weighted sum, a normalization or a small sub-network), so after training it can be enumerated into a truth table of $2^{kb}$ rows~\cite{umuroglu2020logicnets,andronic2024neuralut}; when $kb\le6$, each of its output bits is a single LUT, and its cost depends only on its fan-in, not on the function it learned.
Unrolling instantiates every neuron at every spatial position as its own circuit, wired directly to its inputs, with a register after each layer, so a new image enters every clock cycle and latency is the number of pipeline stages times the clock period.

Let $\mathcal{B}_b=\{0,1\}^b$ denote $b$-bit codes.
The host maps each pixel of an image $X\in\{0,\dots,255\}^{H\times W\times C}$ through a per-channel table to a $b_0$-bit code, $A_0=q_0(X)\in\mathcal{B}_{b_0}^{H\times W\times C}$, folding input normalization and quantization. All the stages after $q_0$ are FPGA logic.
The network is a stack of layers of nodes.
A node $j$ reads a fan-in set $\mathcal{S}_j$ of $k$ activations of $b$ bits each and computes a Boolean function
\begin{equation}
f_j:\mathcal{B}_b^{\,k}\to\mathcal{B}_{b'}, \qquad kb\le 6,
\label{eq:node}
\end{equation}
so it compiles into at most $b'$ LUTs.
Learning the network therefore means learning, for every node, its connectivity $\mathcal{S}_j$ and its function $f_j$.

\begin{figure*}[t]
\centering
\input{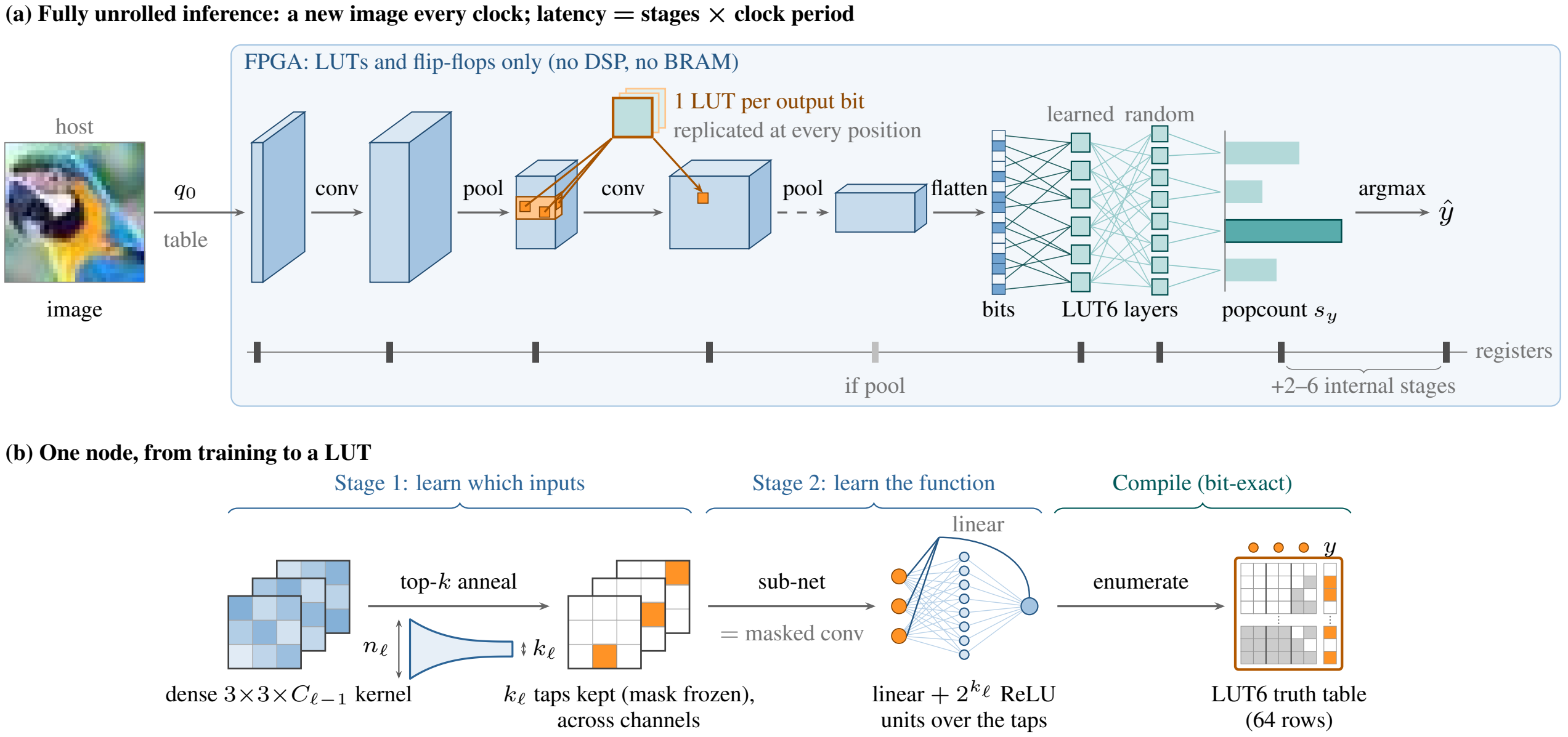}
\caption{PixLUT overview.
(a)~After a host-side table $q_0$ encodes each pixel, the whole classifier is LUTs and registers: $3\times3$ convolutions whose nodes are each one LUT per output bit, replicated at every position, max-pooling, and a weightless LUT head~\cite{bacellar2024dwn} with per-class popcounts $s_y$ and $\hat y=\arg\max_y s_y$.
A new image enters every clock; latency is the 10--15 pipeline stages times the clock period (dashed or grey: only in some designs).
(b)~Each node reads $k_\ell=\lfloor 6/b_{\ell-1}\rfloor$ of the $9C_{\ell-1}$ codes in its window.
Stage~1 learns which ones (annealed top-$k$).
Stage~2 learns the node's function as a linear $+\,2^{k_\ell}$-ReLU sub-network initialised to the masked convolution.
Enumeration then gives a bit-exact LUT6 table.}
\label{fig:overview}
\end{figure*}

As in a CNN, the network splits into a backbone and a head (\cref{fig:overview}).
The backbone learns visual representations. It keeps the image's spatial layout and stacks convolutions, whose kernels are LUT nodes shared across positions, with max-pooling, so that every feature is computed from a local neighbourhood across all channels.
Its output is flattened into a bit vector $z$, and the head, a two-layer DWN~\cite{bacellar2024dwn}, turns these features into class scores. Its first layer of 6-input LUT nodes learns which bits of $z$ each node reads, its second reads the first through fixed random wiring, and the second layer's outputs are grouped by class, so each class score $s_y$ is the popcount of its group and the prediction is $\hat y=\arg\max_y s_y$.
On CIFAR-10, for example, a $32\times32\times3$ image quantized to 3 bits passes through two convolutional layers of 64 and 768 channels, the second with stride 2, each followed by $2\times2$ max-pooling, leaving a $4\times4\times768$ map that is flattened into a 12{,}288-bit vector for the head.
We train our network in 2 stages, first learning the backbone's connectivity and second learning the Boolean functions of its nodes, before bit-exactly compiling the entire network into a fully unrolled circuit.

\subsection{Convolutional LUT backbone}
\label{sec:backbone}

\paragraph{Convolutional LUT nodes.}
To learn from the encoded image as a CNN does, the backbone keeps both the spatial layout and the channel dimension, and layer $\ell$ maps a feature map $A_{\ell-1}$ of $C_{\ell-1}$ channels of $b_{\ell-1}$-bit codes to $A_\ell$ through $3\times3$ convolutions.
A node, one output channel $c$ at one position $p$, sees a window $\mathcal{N}(A_{\ell-1},p)$ of $n_\ell=9C_{\ell-1}$ codes, far more than a LUT's six input bits.
To compile into one LUT per output bit, it may read only
\begin{equation}
k_\ell=\lfloor 6/b_{\ell-1}\rfloor
\label{eq:fanin}
\end{equation}
of them, and these must be chosen across channels as well as positions.
We use $b_0=3$, $b_1=2$ and $b_2=1$, so the first convolution reads two 3-bit codes of the image, the second three 2-bit codes of the first, and the last map is binary.
We therefore learn, for every output channel, a binary mask $M_c\in\{0,1\}^{C_{\ell-1}\times3\times3}$ with $\|M_c\|_0=k_\ell$ over its convolution weights $W_c$, so that layer $\ell$ computes
\begin{equation}
A_\ell[p,c]=q_\ell\Big(\mathrm{BN}_c\big(\langle W_c\odot M_c,\ \mathcal{N}(A_{\ell-1},p)\rangle\big)\Big),
\label{eq:masked}
\end{equation}
where $q_\ell$ is a uniform quantizer to $b_\ell$ bits with a learned scale (the sign for $b_\ell=1$).
The support of $M_c$ is the node's fan-in set $\mathcal{S}_{\ell,c}$ and, as in any convolution, it is shared across positions.

\paragraph{Learning the fan-in.}
Stage~1 learns each $M_c$ with continuous weights by top-$k$ selection, as in top-$k$ spatial selection~\cite{modi2026topk}.
Its goal is the backbone's connectivity, so the backbone trains under a small fully connected surrogate head, which the LUT head replaces once the masks freeze.
Unlike~\cite{modi2026topk}, which selects positions within each two-dimensional kernel, we select over the whole $C_{\ell-1}\times3\times3$ volume, so that a node can combine evidence from different channels.
We also omit its auxiliary regularizer, which concentrates each kernel's energy into $k_\ell$ entries before the mask is applied, as it lowers accuracy here (\cref{tab:ablation}) and annealing the mask alone suffices.
The mask is annealed rather than imposed at once, and at progress $t\in[0,1]$ through the pruning phase, from 20\% to 50\% of Stage~1, it keeps the
\begin{equation}
k_t=k_\ell+(n_\ell-k_\ell)(1-t)^3
\label{eq:anneal}
\end{equation}
largest-magnitude entries of $W_c$~\cite{zhu2017prune}.
The forward pass uses the masked weights, while a straight-through estimator~\cite{bengio2013ste} passes gradients to every entry,
\begin{equation}
\tilde W_c=W_c+\operatorname{sg}\big(W_c\odot M_c-W_c\big),
\label{eq:ste}
\end{equation}
where $\operatorname{sg}$ stops gradients, so a pruned entry can regain magnitude and re-enter the mask.
When $k_t$ reaches $k_\ell$, the mask is frozen and the rest of Stage~1 trains only the selected weights.
The frozen masks fix every node's fan-in; Stage~2 then learns the function each node computes over those inputs.

\paragraph{Learning node functions.}
After the masks freeze, each convolution becomes a set of per-channel functions over its $k_\ell$ selected taps.
A dot product followed by a threshold can express only threshold functions of these inputs, a small fraction of the $2^{2^{k_\ell b_{\ell-1}}}$ functions a LUT can hold~\cite{cover1965geometrical}.
Following NeuraLUT~\cite{andronic2024neuralut}, we add a hidden layer, which lifts this restriction~\cite{BAUM1988}, so that each node becomes a linear residual plus $2^{k_\ell}$ ReLU units over the selected taps $x\in\mathbb{R}^{k_\ell}$,
\begin{equation}
f_{\ell,c}(x)=q_\ell\Big(\mathrm{BN}_c\big(u_c^\top x+v_c^\top\,\mathrm{ReLU}(V_c x+d_c)\big)\Big),
\label{eq:subnet}
\end{equation}
which learns non-linear functions of the selected taps. The extra expressivity costs no hardware because the node is eventually enumerated into a single table.
At the freeze, $u_c$ copies the surviving convolution weights and $v_c=0$, so the sub-network starts as the masked convolution exactly.
Replacing it by the linear residual alone costs 0.4 points on CIFAR-10 (\cref{tab:ablation}).
Stage~2 trains the sub-networks and the head jointly, and we evaluate an exponential moving average of its weights.

\begin{algorithm}[t]
\caption{Training and compiling PixLUT.}
\label{alg:train}
\footnotesize
\begin{algorithmic}[1]
\Require fan-ins $k_\ell$ and code widths $b_\ell$ of every layer
\Statex \textbf{Stage 1} (connectivity), with dense weights $W$ and a small quantized surrogate head
\For{each epoch}
  \State $k_t\gets$ annealing schedule (\cref{eq:anneal})
  \State $M_c\gets$ top-$k_t$ entries of $|W_c|$, for every output channel $c$
  \State train $W$ through the masked weights (\cref{eq:masked,eq:ste})
\EndFor \Comment{$M_c$ is frozen once $k_t=k_\ell$}
\Statex \textbf{Stage 2} (functions)
\State replace every node by its sub-network (\cref{eq:subnet}), initialised to the masked convolution
\State replace the surrogate head by the LUT head
\State train backbone and head jointly
\Statex \textbf{Compile}
\State enumerate every node into a table (\cref{eq:enum})
\State emit the unrolled netlist and verify it on the evaluation split
\end{algorithmic}
\end{algorithm}

\subsection{Bit-exact compilation}
\label{sec:compile}
After Stage~2, every node is a fixed function of a few codes, so we compile it by evaluating it on every input it can receive.

\paragraph{Backbone.}
Let $\lambda_\ell:\mathcal{B}_{b_\ell}\to\mathbb{R}$ map each code to the level that $q_\ell$ assigns it.
Output channel $c$ of layer $\ell$ becomes the table
\begin{equation}
T_{\ell,c}(a)=\lambda_\ell^{-1}\big(f_{\ell,c}(\lambda_{\ell-1}(a))\big),\quad a\in\mathcal{B}_{b_{\ell-1}}^{\,k_\ell},
\label{eq:enum}
\end{equation}
with $\lambda_{\ell-1}$ applied to each of the $k_\ell$ codes, into which $\mathrm{BN}_c$ and $q_\ell$ of \cref{eq:subnet} fold at no hardware cost.
We evaluate $f_{\ell,c}$ with the model's own modules, so each table is the trained function exactly, and max-pooling is a maximum over codes, an OR on 1-bit maps.

\paragraph{Head and netlist.}
The head needs no enumeration, as its nodes are already hard lookups of six bits~\cite{bacellar2024dwn}: its tables are the signs of the learned entries, and its wiring is the argmax of the learned mapping in the first layer and the fixed random one in the second.
Each $s_y$ is the popcount of its group without the group sum's positive temperature, which does not change the argmax, and the circuit registers the lowest maximal index as $\hat y$, as the trained model does.
Instantiating every table at each of its positions gives a netlist of native FPGA logic only, LUTs and flip-flops with carry chains in a few adders and no DSPs or block RAM, and every table is fully specified, so the circuit matches the model on every input, not only those seen in training.

Each node is one LUT per output bit at every position, so the circuit's analytical size is known before synthesis,
\begin{equation}
N_{\mathrm{LUT}}=\sum_{\ell}H_\ell W_\ell C_\ell b_\ell+N_1+N_2+N_{\mathrm{pool}}+N_{\mathrm{sum}},
\label{eq:cost}
\end{equation}
with $N_1$, $N_2$ the head's layers and $N_{\mathrm{pool}}$, $N_{\mathrm{sum}}$ the pooling and the popcounts with argmax.
Synthesis only shrinks it, removing 20\% and 24\% of the two convolutions' LUTs in the CIFAR-10 design, where the first convolution and its pooling are 55\% of the circuit.

For every PixLUT design in \cref{tab:main}, a cycle-accurate Verilator~\cite{snyder2026verilator} simulation of the emitted Verilog, fed one image per clock, reproduces the trained model's prediction on every image of the evaluation split (\cref{sec:experiments}) at a single latency, the cycle count we report, so the bit-exact accuracy of each routed design is that of the circuit.

\begin{figure}[t]
\centering
\includegraphics[width=\linewidth]{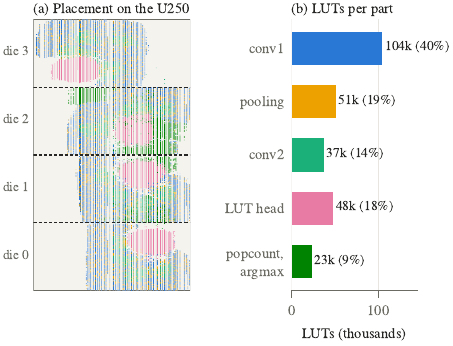}
\caption{The routed CIFAR-10 design on the U250.
(a)~Every placed LUT, coloured by the part of the network it implements, with colours as in (b); dashed lines are the die boundaries.
Each die holds one stripe of the backbone and one block of the head.
(b)~LUTs per part.}
\label{fig:floorplan}
\end{figure}

\subsection{Routing on a multi-die FPGA}
\label{sec:scale}
Unrolled, every LUT input is its own wire, and the U250's four dies~\cite{amd2023ds962} are joined by only 23{,}040 wires at each boundary, which the dense layers of the larger baselines exceed (RQ3, \cref{sec:experiments}).
A $3\times3$ convolution reads only neighbouring positions, so with image stripe $d$ of every convolution placed on die $d$, only the rows at the edge of each stripe need wires to the next die.
The head is the only part whose learned wiring could reach anywhere in the feature map, and this all-to-all wiring is what keeps the larger baselines from routing, so we bound its reach.
The last feature map is flattened position by position and split into $B$ contiguous chunks, and the head into $B$ blocks, each placed with its stripe (\cref{fig:floorplan}); a first-layer node learns its six inputs only within its block's chunk, and a second-layer node reads only first-layer nodes of its block.
Each class's group holds nodes of every block, so every class still sees the whole image.
Class scores are summed per die before they cross, and the popcount and argmax are pipelined, for 10--15 cycles in all (details in \cref{sec:supp-routing}).
On CIFAR-10, heads of 16k and 32k first-layer nodes do not route as a single block, and the four-block design does not route without the per-die placement.
Splitting the head into blocks costs 0.45--1.20 points (\cref{tab:ablation}).
With both measures, our CIFAR-10 to ImageNet designs, of 264.0k--695.5k LUTs, close timing at 200--204\,MHz.
In these designs, the busiest die boundary uses 32--74\% of its wires, and all but six of the 143{,}100 convolution LUTs of the CIFAR-10 design lie on the die of their stripe (\cref{fig:routing,tab:paths}).

\section{Experiments}
\label{sec:experiments}

\paragraph{Setup.}
Every design for which we report hardware, except the cited FPGN-MLP, is placed and routed on an AMD Alveo U250 (xcu250-figd2104-2L-e) with the same out-of-context Vivado 2025.2 flow.
LUTs and FFs are post-route counts; latency is cycles $\times$ clock period, with a new image every clock and the class index registered at the output.
MNIST~\cite{lecun2010mnist}, Fashion-MNIST~\cite{xiao2017fashion} and CIFAR~\cite{krizhevsky2009learning} use the test split; Tiny ImageNet~\cite{le2015tiny} and ImageNet~\cite{deng2009imagenet} use the 10k and 50k validation splits, for which we also report top-5 from the netlist's bit-exact class scores (the circuit itself outputs only the top-1 class), breaking ties toward the lower class index as the hardware argmax does.
Following the PolyLUT and NeuraLUT code bases~\cite{andronic2023polylut,andronic2024neuralut}, we hold out no validation split, so for PixLUT we report the final-epoch accuracy, with no checkpoint selection, as the mean and standard deviation over three seeds.
Its hardware is the seed-42 design at the stage-2 checkpoint with the best test accuracy, evaluated every five epochs (bit-exact accuracy in \cref{tab:main}); every baseline we train is one run selected by the same rule, and the baselines' seed runs vary by less than one point.

\paragraph{Baselines.}
We compare with five flat networks, DWN~\cite{bacellar2024dwn}, LILogicNet~\cite{fojcik2025lilogic}, TreeLUT~\cite{khataei2025treelut}, NeuraLUT-Assemble~\cite{andronic2025assemble} and KANEL\'E~\cite{hoang2026kanele}, and two convolutional logic-gate networks, LogicTreeNet~\cite{petersen2024convdlgn} and CompactLogic~\cite{wang2026compactlogic}.
Each runs from its official code, with its published recipe where it has one and its model size swept as ours, except on ImageNet (one size each); LogicTreeNet has no official code, so we train the reimplementation that CompactLogic uses, and BitLogic~\cite{buhrer2026bitlogic} and FPGN-MLP~\cite{liang2026fpgn} are cited from their papers.
Our runs come within 0.6 points of each method's published result, except NeuraLUT-Assemble and LogicTreeNet on MNIST (0.9 and 0.7 points below).
We check every baseline netlist against its own model on the full test split.
Each one ends in our registered argmax, so every design we route outputs the class index.
This added stage, the different device and our clock search are why our baseline latencies differ from the published ones.
All designs go through the same clock search, and the most accurate model of each method that closes timing is reported; \cref{sec:supp-baselines} details the reproductions, our changes to upstream code and the routing procedure.

\paragraph{Research questions.}
We ask which parts of a convolutional LUT network matter (RQ1), whether it beats flat LUT and logic-gate networks in the same one-image-per-clock regime (RQ2), whether it scales to many classes (RQ3), and what its accuracy costs in LUTs and latency (RQ4).

\begin{figure}[t]
\centering
\includegraphics[width=\linewidth]{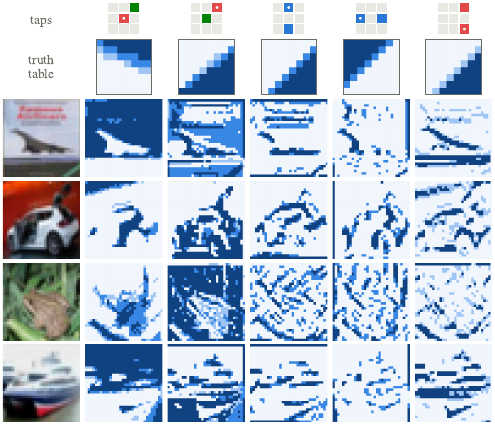}
\caption{What five of the 64 first-layer nodes learn on CIFAR-10.
Taps: the two pixels a node reads in its $3\times3$ window, coloured by input channel (red, green, blue).
Truth table: its 2-bit output, from 0 (light) to 3 (dark), over the 3-bit codes of the dotted tap (vertical) and the other (horizontal).
Below: the node's output at every position of four test images.}
\label{fig:conv1}
\end{figure}

\paragraph{RQ1: Which parts of PixLUT matter.}
\Cref{tab:ablation} changes one thing at a time from a CIFAR-10 reference, the backbone of PixLUT with an $8000\to4000$ head in one block (73.46\%, 194.6k LUTs).
Annealing matters most in training, since a hard top-$k$ cut at the freeze costs 4.84 points.
Starting each sub-network at the masked convolution adds 2.88 points, the surrogate head of Stage~1 1.40, and the sub-network's hidden layer 0.42 at no hardware cost, since each node is one table either way.
Random fixed taps lose 2.08 points and need 10\% more LUTs than learned taps, and restricting each node's taps to one input channel costs 4.80 points.
The rank-decay penalty of~\cite{modi2026topk} lowers accuracy by 0.83 points, so we omit it.
In the head, learned layer-1 wiring adds 6.87 points, and random wiring does not route; a funnel LUT tree and a TT-Sparse head lose 11.68 and 28.06 points, the latter at 64.8k LUTs because Vivado removes the backbone logic its sparse head never reads.
On CIFAR-100, the design of \cref{tab:main} with random fixed taps reaches 47.35\%, against 49.37\% with learned taps.
A second convolution adds 4.38 points over one, and a third does not help (73.57\%) at 50.1k more LUTs.
\Cref{sec:supp-ablation} varies the width and fan-in of the second convolution; $k_2=3$ is the largest fan-in that fits one LUT (\cref{eq:fanin}), a fourth tap adds 0.21 points for 55.8k more LUTs, and $k_2=6$ with 1-bit codes costs 0.77 points but saves 56.8k LUTs.
The head matters as much as depth, as $4000\to2000$ costs 2.42 points, and one-block heads of $16000\to8000$ and $32000\to16000$ gain 1.67 and 3.22 points but do not route.
Block-local heads route at a cost of 0.45 and 1.20 points at 16k (2 and 4 blocks) and 0.85 at 32k (4 blocks), the CIFAR-10 design of PixLUT (\cref{sec:scale}).

\begin{table}[t]
\centering
\footnotesize
\caption{CIFAR-10 ablations, one change each from the reference (conv2 768 channels, $k_2=3$, head $8000\to4000$ in one block), seed 42.
LUTs of the routed design; for the designs that do not route ($\times$), after placement.
Routes: $\checkmark$ placed and routed at 8.0\,ns without clock search (the penalty row at 5.0\,ns, the last row at 4.9\,ns); $\times$ does not route.
Blocked heads also use the position-major bus (\cref{sec:scale}); the last row is the CIFAR-10 design of \cref{tab:main}, and \cref{tab:ablation-full} adds the width and fan-in of the second convolution.}
\label{tab:ablation}
\setlength{\tabcolsep}{3pt}
\begin{tabular}{lrrr}
\toprule
Change & Acc.\ (\%) & LUTs & Routes \\
\midrule
Reference                                 & 73.46 & 194.6k & $\checkmark$ \\
\midrule
\multicolumn{4}{l}{\emph{Training}} \\
Hard top-$k$ cut (no annealing)           & 68.62 & 178.5k & $\checkmark$ \\
No S1 surrogate head                      & 72.06 & 189.4k & $\checkmark$ \\
Linear nodes                              & 73.04 & 189.2k & $\checkmark$ \\
Random sub-net init                       & 70.58 & 174.6k & $\checkmark$ \\
Random fixed taps                         & 71.38 & 214.5k & $\checkmark$ \\
Taps from one input channel               & 68.66 & 171.3k & $\checkmark$ \\
With rank-decay penalty~\cite{modi2026topk} & 72.63 & 186.8k & $\checkmark$ \\
\midrule
\multicolumn{4}{l}{\emph{Head}} \\
Random layer-1 wiring                     & 66.59 & 212.9k & $\times$ \\
Funnel LUT-tree head                      & 61.78 & 265.7k & $\checkmark$ \\
TT-Sparse head~\cite{soegeng2026ttsparse} & 45.40 & 64.8k  & $\checkmark$ \\
\midrule
\multicolumn{4}{l}{\emph{Architecture}} \\
Depth 1 (conv1 only)                      & 69.08 & 62.3k  & $\checkmark$ \\
Depth 3 (+ conv3, 768 ch)                 & 73.57 & 244.7k & $\checkmark$ \\
Head $4000\to2000$                        & 71.04 & 168.3k & $\checkmark$ \\
Head $16000\to8000$, 1 block              & 75.13 & 218.3k & $\times$ \\
Head $16000\to8000$, 2 blocks             & 74.68 & 211.1k & $\checkmark$ \\
Head $16000\to8000$, 4 blocks             & 73.93 & 210.6k & $\checkmark$ \\
Head $32000\to16000$, 1 block             & 76.68 & 250.8k & $\times$ \\
Head $32000\to16000$, 4 blocks            & 75.83 & 264.0k & $\checkmark$ \\
\bottomrule
\end{tabular}
\end{table}

\begin{table*}[t]
\centering
\caption{Comparison with LUT and logic-gate networks on one FPGA (AMD Alveo U250, same flow).
Top-1 accuracy (\%), with top-5 in parentheses for Tiny ImageNet and ImageNet.
PixLUT accuracy is the final-epoch mean $\pm$ s.d.\ over three seeds; its hardware and top-5 are those of the routed seed-42 design, whose bit-exact top-1 is 98.39 and 99.24\% (MNIST), 91.20 and 92.19\% (F-MNIST), 75.83, 49.37, 27.54 and 11.01\%.
Best placed-and-routed accuracy in bold.}
\label{tab:main}
\setlength{\tabcolsep}{3pt}\renewcommand{\arraystretch}{0.89}
\begin{tabular}{llrrrrrr}
\toprule
Dataset & Method & Acc.\ (\%) & LUTs & FFs & $F_{\max}$ (MHz) & Clock (ns) $\times$ cycles & Latency (ns) \\
\midrule
\multirow{10}{*}{MNIST}
 & DWN~\cite{bacellar2024dwn}                      & 98.23 & 4.2k   & 4.7k   & 385 & 2.6 $\times$ 7   & 18.2 \\
 & LILogicNet~\cite{fojcik2025lilogic}             & 98.90 & 45.4k  & 31.5k  & 189 & 5.3 $\times$ 7   & 37.1 \\
 & TreeLUT~\cite{khataei2025treelut}               & 96.58 & 4.2k   & 2.6k   & 286 & 3.5 $\times$ 5   & 17.5 \\
 & NeuraLUT-A~\cite{andronic2025assemble}   & 97.00 & 5.1k   & 5.4k   & 435 & 2.3 $\times$ 8   & 18.4 \\
 & KANEL\'E~\cite{hoang2026kanele}                 & 96.22 & 4.2k   & 4.5k   & 417 & 2.4 $\times$ 11  & 26.4 \\
 & LogicTreeNet~\cite{petersen2024convdlgn}$^\dagger$ & 98.52 & 56.1k & 0.9k & 80 & 12.5 $\times$ 4 & 50.0 \\
 & CompactLogic~\cite{wang2026compactlogic}$^\ddagger$ & 99.38 & --     & --     & --  & --               & --   \\
 & BitLogic~\cite{buhrer2026bitlogic}$^\S$         & 97.84 & --     & --     & --  & --               & --   \\
 & PixLUT (small)                                   & 98.21\,{\scriptsize$\pm$\,0.02} & 17.1k  & 16.5k  & 556 & 1.8 $\times$ 10  & 18.0 \\
 & PixLUT                                           & \textbf{99.14}\,{\scriptsize$\pm$\,0.24} & 270.7k & 245.1k & 208 & 4.8 $\times$ 11 & 52.8 \\
\midrule
\multirow{10}{*}{F-MNIST}
 & DWN                                   & 90.59 & 35.2k  & 32.0k  & 169 & 5.9 $\times$ 7   & 41.3 \\
 & LILogicNet                                      & 90.26 & 91.0k  & 66.2k  & 132 & 7.6 $\times$ 7   & 53.2 \\
 & TreeLUT                                         & 90.28 & 86.9k  & 17.5k  & 89  & 11.2 $\times$ 5  & 56.0 \\  %
 & NeuraLUT-A                               & 86.92 & 6.5k   & 7.2k   & 455 & 2.2 $\times$ 8   & 17.6 \\
 & KANEL\'E                                        & 77.28 & 6.2k   & 6.8k   & 323 & 3.1 $\times$ 12  & 37.2 \\ %
 & LogicTreeNet$^\dagger$                          & 86.11 & 51.9k  & 0.9k   & 81  & 12.3 $\times$ 4  & 49.2 \\
 & CompactLogic$^\ddagger$                         & 86.41 & --     & --     & --  & --               & --   \\
 & BitLogic$^\S$                                   & 89.16 & --     & --     & --  & --               & --   \\
 & PixLUT (small)                                   & 91.19\,{\scriptsize$\pm$\,0.09} & 39.9k  & 37.7k  & 345 & 2.9 $\times$ 10  & 29.0 \\
 & PixLUT                                           & \textbf{91.80}\,{\scriptsize$\pm$\,0.10} & 94.7k & 82.9k & 286 & 3.5 $\times$ 10 & 35.0 \\
\midrule
\multirow{9}{*}{CIFAR-10}
 & DWN                                   & 61.92 & 38.1k  & 46.7k  & 100 & 10.0 $\times$ 6  & 60.0 \\
 & LILogicNet                                      & 57.81 & 154.5k & 101.1k & 145 & 6.9 $\times$ 6   & 41.4 \\
 & TreeLUT                                         & 53.52 & 61.8k  & 30.5k  & 145 & 6.9 $\times$ 5   & 34.5 \\
 & NeuraLUT-A                               & 44.22 & 4.7k   & 7.5k   & 455 & 2.2 $\times$ 8   & 17.6 \\
 & LogicTreeNet$^\dagger$                          & 59.99 & 61.7k & 9.3k & 69 & 14.5 $\times$ 4 & 58.0 \\
 & CompactLogic$^\ddagger$                         & 72.56 & --     & --     & --  & --               & --   \\
 & FPGN-MLP~\cite{liang2026fpgn}$^\P$             & 58.1  & 15.3k  & 2.6k   & 353 & 2.8 $\times$ 2   & 5.7  \\
 & BitLogic$^\S$                                   & 58.06 & --     & --     & --  & --               & --   \\
 & PixLUT                                           & \textbf{74.87}\,{\scriptsize$\pm$\,0.40} & 264.0k & 244.3k & 204 & 4.9 $\times$ 12 & 58.8 \\
\midrule
\multirow{8}{*}{CIFAR-100}
 & DWN                                   & 30.46 & 76.4k  & 70.4k  & 67  & 15.0 $\times$ 7  & 105.0 \\
 & LILogicNet                                      & 27.88 & 150.0k & 105.3k & 139 & 7.2 $\times$ 7   & 50.4 \\
 & TreeLUT                              & 21.38 & 91.9k  & 30.7k  & 80  & 12.5 $\times$ 6  & 75.0 \\
 & NeuraLUT-A                              & 14.95 & 24.5k  & 26.0k  & 263 & 3.8 $\times$ 9   & 34.2 \\
 & LogicTreeNet$^\dagger$                          & 30.88 & --     & --     & --  & --               & --   \\
 & CompactLogic$^\ddagger$                         & 38.08 & --     & --     & --  & --               & --   \\
 & BitLogic$^\S$                                   & 18.82 & --     & --     & --  & --               & --   \\
 & PixLUT                                           & \textbf{48.62}\,{\scriptsize$\pm$\,0.53} & 664.9k & 524.7k & 200 & 5.0 $\times$ 13 & 65.0 \\
\midrule
\multirow{5}{*}{Tiny ImageNet}
 & DWN                                   & 12.55\topfive{30.48} & 76.2k  & 70.8k  & 78  & 12.9 $\times$ 8  & 103.2 \\
 & LILogicNet                                      & 11.96\topfive{27.80} & 146.4k & 104.8k & 135 & 7.4 $\times$ 8   & 59.2 \\
 & LogicTreeNet$^\dagger$                          & 14.96\topfive{34.72} & --     & --     & --  & --               & --   \\
 & CompactLogic$^\ddagger$                         & 19.19\topfive{39.69} & --     & --     & --  & --               & --   \\
 & PixLUT                                           & \textbf{27.21}\,{\scriptsize$\pm$\,0.11}\topfive{50.42} & 638.2k & 518.7k & 204 & 4.9 $\times$ 13 & 63.7 \\
\midrule
\multirow{3}{*}{ImageNet}
 & DWN                                             & 1.79\topfive{6.61}  & 42.8k  & 46.9k  & 182 & 5.5 $\times$ 8   & 44.0 \\
 & LILogicNet                                      & 2.96\topfive{8.18}  & 129.3k & 97.5k  & 135 & 7.4 $\times$ 8   & 59.2 \\
 & PixLUT                                           & \textbf{11.07}\,{\scriptsize$\pm$\,0.15}\topfive{24.39} & 695.5k & 579.1k & 200 & 5.0 $\times$ 15 & 75.0 \\
\bottomrule
\end{tabular}
\par\vspace{3pt}
\begin{minipage}{\linewidth}\footnotesize
$^\S$Cited software accuracy (code not released), not implemented.
$^\dagger$Reproduced from \url{https://github.com/eth-sri/CompactLogic} (no official code); its CIFAR-100 and Tiny ImageNet models do not place or route on the U250.
$^\ddagger$Authors' Verilog, unchanged, fails placement or routing on the U250.
$^\P$Cited, not re-implemented; hardware does not include the argmax.
\end{minipage}
\end{table*}

\paragraph{What the first layer learns.}
A first-layer node reads two pixels of eight levels each, so its 64-entry truth table is its whole function, and \cref{fig:conv1} shows it next to the node's response on test images.
Most tables are a triangle on one side of the diagonal, so the output depends on which of the two taps is brighter.
When the taps are two positions of one colour, this is an edge detector, and its maps trace horizontal or vertical contours (columns 3--5).
When they are two colours, it measures colour contrast and separates an object from its background (column 2).
A few nodes follow the brightness of a single tap instead (column 1).
Of the 64 nodes, 40 read two positions of one colour, 8 two colours at one position and 14 both, and about two thirds of the tables are such comparisons.
Without being told to, the layer learns the edge and colour-opponent filters familiar from the first layer of a CNN, each as a single lookup.

\paragraph{RQ2: Accuracy in the same regime.}
PixLUT is the most accurate placed-and-routed design in this comparison on all six datasets (\cref{tab:main}).
At the final epoch and averaged over three seeds, the margin on MNIST and Fashion-MNIST is small, 0.24 points over LILogicNet and 1.21 over DWN; from CIFAR-10 on, PixLUT leads the best routed baseline by 13.0, 18.2, 14.7 and 8.1 points, or 1.2, 1.6, 2.2 and 3.7$\times$ its accuracy.
Convolution alone does not explain the margin: on CIFAR-10 the convolutional logic-gate networks, LogicTreeNet (59.99\%, routed) and CompactLogic (72.56\%, not placeable), stay below PixLUT, as do the flat networks, which lie between 44.22 and 61.92\%.
Of the accuracies in \cref{tab:main} without hardware, only CompactLogic's 99.38\% on MNIST exceeds ours (99.14\%).

\paragraph{RQ3: Scaling with the number of classes.}
The ratio to the best routed baseline grows with the number of classes, because the baselines fall toward chance faster than PixLUT.
On ImageNet, each baseline is trained at a single size.
The baselines' larger models are more accurate, but none of them trains, synthesizes, places or routes on our hardware with any aid we tried (\cref{sec:supp-stop}).
On datasets from CIFAR-10 and bigger, none of these larger models reaches PixLUT even before routing.
The closest is CompactLogic's 72.56\% model on CIFAR-10, which synthesizes to 1.35M LUTs and needs 546{,}795 nets across one die boundary, which is too congested for routing on device. 
The local wiring of PixLUT keeps its 264.0k--695.5k LUTs designs within the die boundary limit (\cref{sec:scale}).

\paragraph{RQ4: Cost in LUTs and latency.}
The accuracy costs LUTs rather than latency: from CIFAR-10 on, PixLUT uses 264.0k--695.5k LUTs against at most 154.5k for any baseline we route, yet its latency lies within theirs, 58.8\,ns on CIFAR-10 against 17.6--60.0\,ns, and 65.0 and 63.7\,ns on CIFAR-100 and Tiny ImageNet against 34.2--105.0 and 59.2--103.2\,ns; on ImageNet both baselines are faster (44.0 and 59.2 \vs 75.0\,ns).
On MNIST, all baselines are faster than our most accurate design (52.8\,ns), and our 17.1k-LUT design matches DWN at its latency (98.21 \vs 98.23\% in 18.0 \vs 18.2\,ns) with 4.1$\times$ its LUTs.

\section{Limitations}
\label{sec:limitations}
As each class score is the popcount of a group of head nodes, the head needs to scale according to the number of classes to maintain the granularity of the scores. However, this also scales with the LUTs and more importantly, the wiring that cross between dies, which may produce congestion during routing. Thus, this motivates future work to find a head or framework whose cost grows more slowly with the number of classes to improve on large label sets.

\section{Conclusion}
\label{sec:conclusion}
We presented PixLUT, a convolutional network built entirely from 6-input LUTs, with a convolutional backbone and a weightless LUT head that scores each class by a popcount.
Training first learns which few inputs each backbone neuron reads, across channels and positions, and then a nonlinear function of them, after which the network compiles exactly into a fully unrolled circuit of native FPGA logic.
Keeping the wiring local lets designs of up to 696k LUTs route on one multi-die FPGA.
At one image per clock, PixLUT is more accurate than the LUT and logic-gate networks we compared against on every dataset, and the gap grows with the difficulty of the task.
These results show that convolutional structure can be learned and kept in a fully unrolled LUT circuit without giving up nanosecond latency.

Building on this work, we identify a directions for future research. Because each table absorbs any choice of quantization levels at no hardware cost, learned non-uniform quantizers are natural next steps which can improve this framework. The unrolled backbone also computes a feature at every position, which suits dense prediction like segmentation, where each position would output its own label.

\section*{LLM Usage Statement}
We used a large language model assistant to help write and refactor code, to launch and monitor training, synthesis and simulation runs, to check the numbers in this paper against our result logs, and to draft and edit text.
The authors designed the method and the experiments, verified the results, and take full responsibility for the content of this paper.

{
    \small
    \bibliographystyle{ieeenat_fullname}
    \bibliography{main}
}

\clearpage
\appendix
\setcounter{table}{0}\renewcommand{\thetable}{S\arabic{table}}
\setcounter{figure}{0}\renewcommand{\thefigure}{S\arabic{figure}}
\setcounter{equation}{0}\renewcommand{\theequation}{S\arabic{equation}}
\begin{center}{\Large\bf Appendix}\end{center}

\section{Architectures and per-run configurations}
\label{sec:arch}
Every design has two $3\times3$ convolutions, each followed by $2\times2$ max-pooling (the second pool is omitted where marked), and a two-layer DWN head~\cite{bacellar2024dwn} whose layer-2 nodes are grouped by class and summed (popcount) into class scores.
Inputs are 3-bit codes ($k_1=\lfloor6/3\rfloor=2$), conv1 emits $b_1$-bit codes ($k_2=\lfloor6/b_1\rfloor$) and conv2 emits 1-bit maps that are flattened into the head's input bus.
\Cref{tab:arch} gives every setting that differs between the designs of \cref{tab:main} and the CIFAR-10 reference of Tab.~2; \cref{tab:common} gives the settings they share, and \cref{tab:abl} the change made by each ablation of Tab.~2.
The CIFAR-100, Tiny ImageNet and ImageNet designs read up to $9\times80=720$ codes per conv2 node.

\begin{table*}[t]
\centering\scriptsize
\setlength{\tabcolsep}{2.6pt}
\caption{Architecture and training configuration of every design of \cref{tab:main} and of the CIFAR-10 reference of Tab.~2.
$b_1$: bits of conv1's codes; $k_1$, $k_2$: fan-in of conv1 and conv2 nodes (codes); $C_1$, $C_2$: channels; $s_1$, $s_2$: strides; pool 2: second max-pool; bus: bits into the head and their order (channel-major chw or position-major hwc); head: DWN layer-1 $\to$ layer-2 nodes; blocks: block-local groups of layer 1 / layer 2; $\tau$: group-sum temperature in training; S1 head: hidden width of the Stage-1 surrogate head; sub-net: hidden width of each conv2 node's sub-network (\cref{eq:subnet}; conv1 nodes: 4); epochs: Stage 1 / Stage 2.}
\label{tab:arch}
\begin{tabular}{llcccrrccrlcrrcrc}
\toprule
Dataset & Design & $b_1$ & $k_1$ & $k_2$ & $C_1$ & $C_2$ & $s_1$/$s_2$ & pool 2 & bus (bits) & head & blocks & $\tau$ & S1 head & sub-net & epochs & MixUp \\
\midrule
MNIST & small & 2 & 2 & 3 & 12 & 96 & 2/1 & -- & 4{,}704 chw & $4000\to2000$ & 1/1 & 14.14 & 64 & 8 & 100/200 & 0.2 \\
MNIST & main & 2 & 2 & 3 & 64 & 768 & 1/1 & \checkmark & 37{,}632 hwc & $40000\to20000$ & 16/400 & 44.72 & 64 & 8 & 100/200 & 0.2 \\
F-MNIST & small & 1 & 2 & 6 & 16 & 192 & 1/2 & -- & 9{,}408 chw & $8000\to4000$ & 1/1 & 20.00 & 1024 & 64 & 100/200 & -- \\
F-MNIST & main & 2 & 2 & 3 & 32 & 192 & 1/2 & -- & 9{,}408 hwc & $16000\to8000$ & 2/2 & 28.28 & 1024 & 8 & 100/200 & -- \\
CIFAR-10 & reference (Tab.~2) & 2 & 2 & 3 & 64 & 768 & 1/2 & \checkmark & 12{,}288 chw & $8000\to4000$ & 1/1 & 20.00 & 64 & 8 & 400/800 & 0.2 \\
CIFAR-10 & main & 2 & 2 & 3 & 64 & 768 & 1/2 & \checkmark & 12{,}288 hwc & $32000\to16000$ & 4/4 & 40.00 & 64 & 8 & 400/800 & 0.2 \\
CIFAR-100 & main & 2 & 2 & 3 & 80 & 1792 & 1/1 & \checkmark & 114{,}688 hwc & $32000\to128000$ & 16/640 & 17.89 & 1024 & 8 & 400/800 & 0.2 \\
Tiny IN & main & 2 & 2 & 3 & 80 & 1792 & 1/1 & \checkmark & 114{,}688 hwc & $32000\to128000$ & 16/640 & 12.65 & 1024 & 8 & 150/600 & 0.2 \\
ImageNet & main & 2 & 2 & 3 & 80 & 1792 & 1/1 & \checkmark & 114{,}688 hwc & $32000\to128000$ & 16/640 & 5.66 & 1024 & 8 & 48/90 & 0.2 \\
\bottomrule
\end{tabular}
\end{table*}

\begin{table}[t]
\centering\footnotesize
\setlength{\tabcolsep}{3pt}
\caption{Settings shared by every run.}
\label{tab:common}
\begin{tabular}{p{0.34\linewidth}p{0.6\linewidth}}
\toprule
Setting & Value \\
\midrule
Input encoding & per-channel table to 3-bit codes ($q_0$, host side) \\
Stage 1 optimizer & AdamW, lr $10^{-3}$, cosine, weight decay $10^{-4}$ \\
Stage 1 batch & 256 \\
Top-$k$ anneal & cubic (Eq.~(4)), from 20\% to 50\% of Stage~1 \\
Surrogate head & 1-bit weights, 2-bit activations, 2-bit output weights \\
Rank-decay penalty~\cite{modi2026topk} & off ($\beta=0$) \\
Freeze & sub-nets initialised to the masked convolution \\
Stage 2 optimizer & AdamW, lr $2\times10^{-4}$ (backbone), $10^{-2}$ (head), cosine, no weight decay \\
Stage 2 batch & 512 \\
Weight EMA & Stage~2 only, decay 0.9995 \\
DWN head & 6-input nodes, learned layer-1 mapping \\
Augmentation & random crop (pad 4; MNIST, F-MNIST pad 2), horizontal flip (not MNIST) \\
Evaluation & every 5 epochs \\
Seeds & 42, 43, 44 (hardware: seed 42) \\
\bottomrule
\end{tabular}
\end{table}

\begin{table}[t]
\centering\footnotesize
\setlength{\tabcolsep}{3pt}
\caption{The one change of each ablation of \cref{tab:ablation} and \cref{tab:ablation-full}, relative to the CIFAR-10 reference of \cref{tab:arch}.}
\label{tab:abl}
\begin{tabular}{p{0.36\linewidth}p{0.58\linewidth}}
\toprule
Row of Tab.~2 & Change \\
\midrule
Hard top-$k$ cut & no annealing: top-$k$ applied once at the freeze \\
No S1 surrogate head & Stage~1 trains under the DWN head \\
Linear nodes & sub-net without its hidden ReLU layer \\
Random sub-net init & sub-nets randomly initialised at the freeze \\
Random fixed taps & $k_\ell$ random taps per channel, fixed from the first step \\
Taps from one input channel & taps chosen inside one input channel \\
With rank-decay penalty & $\beta=200$, $\alpha=0.05$, warm-up 120 epochs \\
Random layer-1 wiring & DWN layer-1 mapping fixed at random \\
Funnel LUT-tree head & 10{,}240 first-layer head nodes reduced by fan-in-4 LUT funnels (its own configuration, chw bus) \\
TT-Sparse head & every head node a TT-Sparse node \\
Depth 1 & conv1 only, no final pool \\
Depth 3 & + conv3, 768 channels, stride 1 \\
conv2 192 / 384 / 1{,}536 ch & $C_2$ = 192 / 384 / 1{,}536 \\
conv2 384 ch at $16\times16$ & $C_2=384$, $s_2=1$ \\
$k_2=2$ / $k_2=4$ & conv2 fan-in 2 / 4 codes \\
$k_2=6$ (1-bit codes) & $b_1=1$, $k_2=6$ \\
Head $4000\to2000$ & $\tau=14.14$ \\
Head $16000\to8000$, 1/2/4 blocks & $\tau=28.28$; 2 and 4 blocks use the hwc bus \\
Head $32000\to16000$, 1/4 blocks & $\tau=40.00$; 4 blocks use the hwc bus \\
\bottomrule
\end{tabular}
\end{table}

\section{Training configurations}
\label{sec:recipe}
\paragraph{Stage 1.}
The backbone trains with continuous weights, quantized activations and the annealed top-$k$ mask (\cref{eq:anneal}), which starts at 20\% and reaches $k_\ell$ at 50\% of Stage~1, under a surrogate head: a two-layer quantized classifier (1-bit weights, 2-bit hidden activations, 2-bit output weights) whose hidden width is given in \cref{tab:arch}.
It is cheaper to train than the LUT head and is discarded at the freeze.
AdamW, learning rate $10^{-3}$ with a cosine schedule, weight decay $10^{-4}$, batch 256, MixUp ($\alpha=0.2$).
\paragraph{Freeze.}
Each masked convolution becomes per-channel sub-networks over its $k_\ell$ taps (\cref{eq:subnet}, hidden width $2^{k_\ell}$), initialised to the masked convolution, and the surrogate head is replaced by a fresh DWN head.
\paragraph{Stage 2.}
Sub-networks and head train jointly with AdamW (backbone $2\times10^{-4}$, head $10^{-2}$, cosine schedule, no weight decay), batch 512, MixUp ($\alpha=0.2$) and a weight EMA (decay 0.9995); evaluation every five epochs.
\paragraph{Data.}
Random crop (padding 4 on $32\times32$ inputs, 2 on MNIST and Fashion-MNIST) and horizontal flip (not on MNIST); Fashion-MNIST trains without MixUp.
Tiny ImageNet is box-downsampled from $64\times64$, ImageNet-1k from full resolution~\cite{chrabaszcz2017downsampled}.
Every design of \cref{tab:main} is trained with seeds 42, 43 and 44, and its hardware is the seed-42 run; the rank-decay penalty of~\cite{modi2026topk} is off ($\beta=0$).

\section{Compilation details}
\label{sec:supp-compile}
We evaluate $f_{\ell,c}$ with the model's own floating-point modules on the levels each quantizer emits, since a folded BN is exact only in real arithmetic and moves pre-activations on a rounding boundary.
Padding is the code the model pads with in training (the zero level, or $-1$ for a 1-bit feed with levels $\pm1$), so border nodes share the interior table.
Max-pooling follows the quantizer and codes are ordered as their levels, so it is a maximum over codes, an OR on 1-bit maps.

\section{Routing details}
\label{sec:supp-routing}
This section expands Sec.~3.3 and \cref{fig:floorplan}.

\paragraph{Block-local head wiring.}
The backbone is local, so with image stripe $d$ of every convolution pinned to die $d$, only one-row halos cross dies.
Like FPGN's in-order flattening~\cite{liang2026fpgn}, a blocked head reads the last 1-bit map $A_L$ flattened position-major (wiring only), $z_{pC_L+c}=A_L[p,c]$ with $p$ in raster order, so a contiguous chunk of $z$ holds all channels of one stripe, whereas a channel-major chunk spans every die.
We split $z$ into $B$ contiguous chunks and the head into $B$ blocks, each pinned with its stripe: first-layer node $j$ learns its six inputs only within chunk $j \bmod B$, and second-layer nodes draw their random inputs only from first-layer nodes of the same block.
Each class's group holds nodes of every block, so every class still sees the whole image.
On CIFAR-10, one-block heads of 16k and 32k first-layer nodes do not route (45{,}403 nets with routing conflicts at 16k).
Blocked heads of the same sizes route (\cref{tab:ablation}), and the 32k head in four blocks is our CIFAR-10 design.

\paragraph{Per-die reduction.}
Each of the $K$ class scores is reduced die by die,
\begin{equation}
s_y=\sum_{d=1}^{D}\Big[\sum_{i\in\mathcal{G}_{y,d}}h_i(z)\Big],
\label{eq:perdie}
\end{equation}
where $h_i(z)\in\{0,1\}$ is second-layer node $i$, $\mathcal{G}_{y,d}$ the nodes of class $y$ on die $d$, and $D=4$.
Each bracket is summed in registered pieces of at most 16 terms and registered on its die, so only $D$ registered words per class cross dies.
At 4.9\,ns, the four-block CIFAR-10 design does not route without the floorplan (33{,}867 routing errors) and misses by 0.178\,ns with the floorplan but without the per-die sums.
With both, it closes at 4.9\,ns.
The 600k-LUT designs need them too, and the 1{,}000-class design needs one more register on each crossing to close 5.0\,ns; on an earlier training of the same architectures, the class-sum wires took 8.6\,ns at 5.0\,ns, and with $K=1{,}000$ the design still missed by 0.715\,ns without these registers.

\paragraph{Pipelined popcount and argmax.}
Popcount stages are 4--10 LUT levels deep and the argmax tree has $\lceil\log_2 K\rceil$ levels, so we cut the argmax once for 10 classes, twice for 100--200 and three times for 1{,}000.
Popcount pieces, argmax cuts and, where used, per-die and crossing registers add 2--6 stages without changing the function, for 10--15 cycles in all.
With these measures and the baselines' flow (\cref{sec:experiments}), our CIFAR-10 to ImageNet designs, of 264.0k--695.5k LUTs, close timing at 200--204\,MHz.

\begin{figure}[t]
\centering
\includegraphics[width=\linewidth]{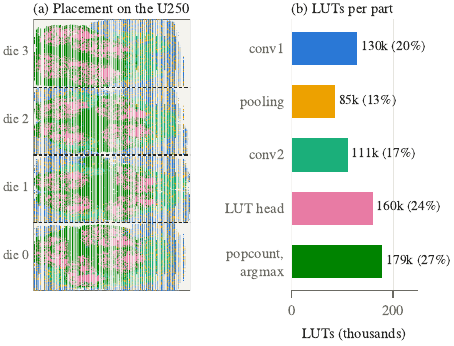}
\caption{The routed CIFAR-100 design on the U250, drawn as \cref{fig:floorplan}.
(a)~Every placed LUT, coloured by the part of the network it implements, with colours as in (b); dashed lines are the die boundaries.
(b)~LUTs per part.}
\label{fig:floorplan-c100}
\end{figure}

\paragraph{Placement at 100 classes.}
\Cref{fig:floorplan-c100} repeats \cref{fig:floorplan} for the CIFAR-100 design, which has 664.9k LUTs and a head of 16 blocks.
The layout is the same at this size.
The backbone spans all four dies, and the first-layer nodes of each head block lie on a single die.
The 16 blocks are spread four to a die.
In the CIFAR-10 design, each die holds one of the four blocks.
What changes is where the LUTs go.
On CIFAR-10 the first convolution and the pooling are 59\% of the circuit, and the head with its popcounts and argmax 27\%.
On CIFAR-100 these shares are 32\% and 51\%, and the popcounts and argmax alone, at 179.2k LUTs, are the largest part.
With 100 classes the cost of the circuit thus moves from the first convolution to the classifier.

\begin{figure*}[t]
\centering
\includegraphics[width=\linewidth]{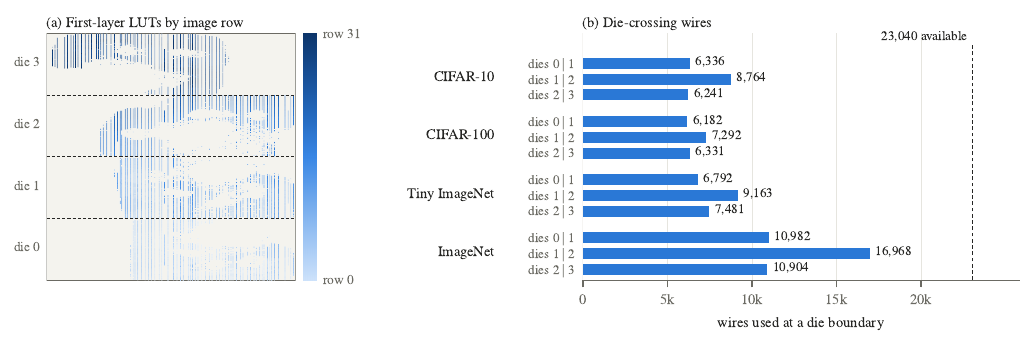}
\caption{(a)~The first-layer LUTs of the routed CIFAR-10 design, coloured by the row of the $32\times32$ image they compute (row 0 is the top of the image); dashed lines are the die boundaries.
(b)~Wires used at each of the three die boundaries by our four largest routed designs, against the 23{,}040 that the device has at each.}
\label{fig:routing}
\end{figure*}

\paragraph{Placement and timing analysis.}
Three measurements on the routed designs show how local the circuit is (\cref{fig:routing,tab:paths}).

\emph{Die crossings.}
We report the wires used at each die boundary, of 23{,}040 available, as provided by Vivado.
At their busiest boundary, our CIFAR-10, CIFAR-100, Tiny ImageNet and ImageNet designs use 8{,}764, 7{,}292, 9{,}163 and 16{,}968 of them, 32--74\% of the supply, with up to 695.5k LUTs on the device.
\Cref{fig:routing}b gives all three boundaries of each design.

\emph{Locality.}
Unrolling gives every image position its own first-layer LUTs, so each LUT belongs to one image row.
\Cref{fig:routing}a colours them by that row, from row 0 at the top of the $32\times32$ image to row 31 at the bottom.
The four bands are the floorplan, which places image rows 0--7, 8--15, 16--23 and 24--31 on dies 0 to 3; all but six of the 143{,}100 convolution LUTs of the CIFAR-10 design lie on the die of their stripe.
Within a die the placer also keeps neighbouring rows together without being told to, so the vertical position of a first-layer LUT follows its image row with a correlation of 0.97.
A $3\times3$ convolution reads only its own row and the two next to it, so these wires stay short and only the rows at the edge of a band cross a die boundary.
The first-layer LUTs of one image position form a cluster, with a spread of one eighth of the layer's horizontally and one thirtieth vertically.
Along a row the placer does not keep image columns in order (correlation 0.13 between image column and horizontal position), so a regular placement in both directions may shorten wires further.

\emph{Timing.}
\Cref{tab:paths} lists the slowest path of each design.
In every design, 65--98\% of the slowest path's delay is wire.
In the three largest designs the slowest path holds a single LUT, so no further register would shorten it.
Since wire sets the clock, the clock depends on how close the placement keeps LUTs that feed each other, and the locality above is what keeps them close.
Without the floorplan, the CIFAR-10 design does not route at all (33{,}867 routing errors).

\begin{table}[t]
\centering
\footnotesize
\caption{Slowest path of each routed PixLUT design: the stage it lies in, the LUTs it passes between two registers, its delay and the share of that delay spent in wires.}
\label{tab:paths}
\setlength{\tabcolsep}{3pt}
\begin{tabular}{llrrr}
\toprule
Design & Stage & LUTs & Delay (ns) & Wire (\%) \\
\midrule
MNIST, small   & popcount                      & 6 & 1.74 & 65 \\
MNIST          & popcount                      & 9 & 4.72 & 78 \\
F-MNIST, small & head $\to$ popcount           & 5 & 2.67 & 82 \\
F-MNIST        & conv2 $\to$ head             & 1 & 3.48 & 95 \\
CIFAR-10       & popcount                      & 7 & 4.84 & 87 \\
CIFAR-100      & input $\to$ conv1             & 1 & 4.97 & 95 \\
Tiny ImageNet  & pooling $\to$ conv2           & 1 & 4.75 & 96 \\
ImageNet       & input $\to$ conv1             & 1 & 4.92 & 98 \\
\bottomrule
\end{tabular}
\end{table}

\section{First-layer truth tables}
\label{sec:supp-tables}
\begin{figure*}[t]
\centering
\includegraphics[width=\linewidth]{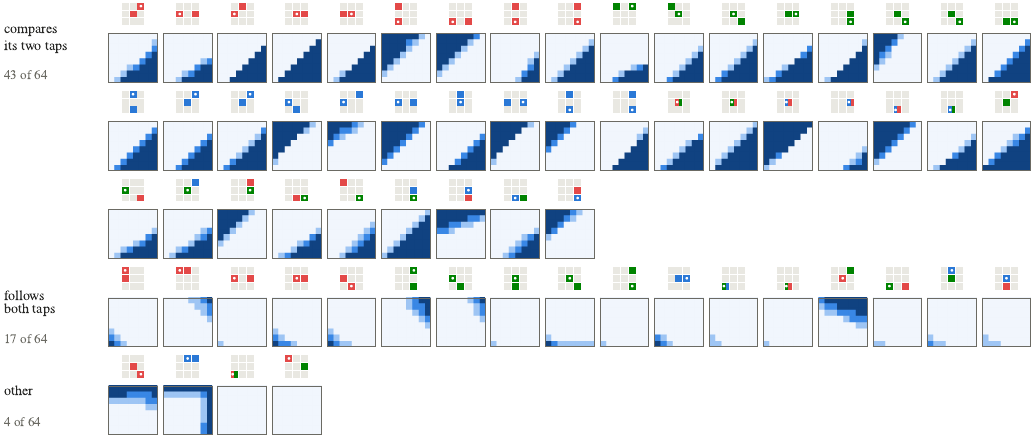}
\caption{All 64 nodes of the first convolution of the CIFAR-10 design.
Top of each tile: the two pixels the node reads in its $3\times3$ window, coloured by input channel (red, green, blue); a split square is two channels of one pixel.
Bottom: its truth table, the 2-bit output from 0 (light) to 3 (dark) over the 3-bit code of the dotted tap (vertical, brighter upwards) and of the other tap (horizontal, brighter to the right).
The last two tables are constant.}
\label{fig:tables}
\end{figure*}
\Cref{fig:tables} shows every node of the first convolution of the CIFAR-10 design, of which \cref{fig:conv1} shows five.
A node reads two 3-bit pixel codes and emits a 2-bit code, so it is two LUTs on the same six inputs and its truth table has $8\times8$ entries.
Each node is reused at every position, so these 64 tables are the whole layer, and they are read from the routed netlist.

A table is read as follows.
Its row is the level of the dotted tap and its column the level of the other, from darkest (bottom, left) to brightest (top, right), so the cells on the diagonal are those where the two pixels are equal.
A table that is dark on one side of the diagonal and light on the other depends on which pixel is brighter and not on how bright the two are, so the node measures contrast and is indifferent to the overall lighting.
It detects an edge when its taps are two neighbouring pixels of one colour, and colour contrast when they are two channels.
The intermediate shades next to the diagonal, where the pixels differ by little, give a weaker output than a large difference, so the node also grades the contrast.
A table that is dark only in a corner responds when both pixels are very dark (bottom left) or very bright (top right).

We group the tables by a property checked entry by entry.
In 43 of them the output never falls as one tap brightens and never rises as the other does, so the node compares its two taps.
In 17 the output moves the same way with both taps, and 13 of these differ from a constant in at most 8 of their 64 entries, the corner tables above.
Two tables are not monotone and two are constant.
By their taps, 40 nodes read two positions of one colour, 9 two colours at one position and 15 both; leaving out the two constant nodes gives the 40, 8 and 14 of \cref{sec:experiments}.

\section{Validation-selected checkpoints}
\label{sec:supp-val}
\begin{table}[!t]
\centering
\footnotesize
\caption{Top-1 test accuracy (\%) with a held-out validation split, seed 42, bit-exact.
Val-sel.: checkpoint chosen on the held-out split.
Test-best: best test accuracy of the same run, logged but not used.
Test-sel.: the seed-42 design of \cref{tab:main}, chosen on test and trained on all training images.
Final: final-epoch mean $\pm$ s.d.\ over three seeds, as in \cref{tab:main}.
Tiny ImageNet is tested on its official validation split.}
\label{tab:val}
\setlength{\tabcolsep}{3pt}
\begin{tabular}{lrrrr}
\toprule
Design & Val-sel. & Test-best & Test-sel. & Final \\
\midrule
MNIST, small        & 98.16 & 98.32 & 98.39 & 98.21\,{\scriptsize$\pm$\,0.02} \\
MNIST               & 99.31 & 99.31 & 99.24 & 99.14\,{\scriptsize$\pm$\,0.24} \\
F-MNIST, small      & 91.05 & 91.10 & 91.20 & 91.19\,{\scriptsize$\pm$\,0.09} \\
F-MNIST             & 91.91 & 92.09 & 92.19 & 91.80\,{\scriptsize$\pm$\,0.10} \\
CIFAR-10            & 74.76 & 75.10 & 75.83 & 74.87\,{\scriptsize$\pm$\,0.40} \\
CIFAR-100           & 48.71 & 48.72 & 49.37 & 48.62\,{\scriptsize$\pm$\,0.53} \\
Tiny ImageNet       & 26.65 & 27.04 & 27.54 & 27.21\,{\scriptsize$\pm$\,0.11} \\
CIFAR-10 reference  & 72.94 & 73.16 & 73.46 & -- \\
\bottomrule
\end{tabular}
\end{table}
The hardware in \cref{tab:main} is built from the seed-42 checkpoint with the best test accuracy (\cref{sec:experiments}).
To see how much this choice adds, we retrained every design of Tab.~3, and the CIFAR-10 reference of Tab.~2, with a validation split held out from the training set: 5{,}000 class-stratified images, or 10{,}000 for Tiny ImageNet.
The Stage-1 hand-off and the Stage-2 checkpoint are both chosen on this split, and the test split is then evaluated once, bit-exact (\cref{tab:val}).

Selecting on test adds little.
Within a run, the validation-selected checkpoint is at most 0.39 points below the best test accuracy, and it stays within 1.07 points of the design in \cref{tab:main}, which was also trained on the held-out images.
On CIFAR-10 it lies 0.11 points below the three-seed final-epoch mean that \cref{tab:main} reports.
On every dataset, the larger design remains above the best routed baseline (98.90, 90.59, 61.92, 30.46 and 12.55\%), even though those baselines keep their test-selected accuracies.
\section{Baseline protocol}
\label{sec:supp-baselines}
Each runs from its official code at a pinned commit in its own environment, and every change to upstream code (datasets, logging, resuming, hardware export) is a separate patch.
LogicTreeNet has no official code, so we train ConvLogic, a community reimplementation archived in CompactLogic's repository, which CompactLogic also uses to reproduce LogicTreeNet.
BitLogic~\cite{buhrer2026bitlogic} links no public code, so we cite its software accuracy; FPGN-MLP~\cite{liang2026fpgn} is cited from its paper.
Before extending a method to a new dataset, we reproduce one of its published results: DWN, LILogicNet, TreeLUT and LogicTreeNet (CIFAR-10, model S) come within 0.5 points (DWN on Fashion-MNIST 0.57 above), and the released checkpoints of CompactLogic and KANEL\'E reproduce their papers.
Two fall short, and we report our numbers: NeuraLUT-Assemble is 1.1 points below on MNIST at its selected epoch (0.9 for its LUT model, 97.00\%) with Brevitas 0.2.1, the first usable release after its pinned one, a gap that follows the Brevitas version (0.28 points with 0.10.2) rather than the seed, and LogicTreeNet's MNIST model reaches 98.52\% against the published 99.23\%.
Each method trains with its published recipe or, where it has none, with our augmentation.
We sweep its model size with a budget like ours, except on ImageNet (one size each), and select its epoch by the same rule.

Where a method emits its own netlist, we use it unchanged (TreeLUT, NeuraLUT-Assemble, KANEL\'E, and CompactLogic's converter for both logic-gate networks); for DWN and LILogicNet, we write it with a register after every layer, as for PixLUT.
Every netlist is checked against the method's own model on the full test split, so a routed design's accuracy is that model's accuracy.
The baselines' own hardware, where it exists, ends at the class scores.
We add our registered argmax (ties toward the lower index), so every design we route ends at the class index.
This extra stage, the different device and our clock procedure are why our baseline latencies differ from the published ones.
Around a netlist that a method emits itself (TreeLUT, NeuraLUT-Assemble, KANEL\'E), whose released hardware ends at the class scores, we add a registered input, the argmax and a registered prediction, three clock cycles for ten classes and more where the argmax tree is cut again for 100 or more classes.
On MNIST, TreeLUT thus goes from about 2 cycles in its paper (its latency times its clock frequency) to 5 in \cref{tab:main}, and KANEL\'E from about 8 to 11.
The netlists we write ourselves (DWN, LILogicNet and PixLUT) end in the same registered argmax, so the cycle counts of Tab.~3 are like for like.
Each design is routed at 12.5\,ns, or at looser periods until one closes, and once more at the period implied by that run's slack, rounded up to 0.1\,ns, and the tighter period that closes is reported.
Our designs follow the same rule, starting from the period at which an earlier build of the design closed.
A baseline that does not route gets the aids of our designs (column density cap, both placement directives) and more (spreading directives, looser caps and, last, 15\,ns), and the most accurate model of each method that closes timing is reported.

\section{Larger baseline models}
\label{sec:supp-stop}
Their larger models are more accurate but do not place or route on the U250 with any aid we tried: on CIFAR-10, DWN with a 32k-node first layer (63.35\%), LILogicNet with 128k gates (60.73\%) and TreeLUT with 3{,}000 trees (57.42\%), and on Tiny ImageNet, DWN's largest model (13.66\%).
LogicTreeNet on CIFAR-100 and Tiny ImageNet and every CompactLogic model fail in the same way, including CompactLogic's 72.56\% CIFAR-10 model, which synthesizes to 1.35M LUTs and needs 546{,}795 nets across one die boundary.
CompactLogic's paper reports gate counts and no placed-and-routed design.
The dense layers of these models need more wires between dies than the device has.
The local wiring of PixLUT keeps its 264.0k--695.5k-LUT designs within that limit (\cref{sec:scale}).
Other baselines stop earlier: NeuraLUT-Assemble exceeds a 141\,GB GPU in training on Tiny ImageNet, TreeLUT's larger CIFAR-100 models exceed 160\,GB in synthesis, and KANEL\'E, whose first layer would need pruning to about 1--2\% of its edges on 3{,}072 or more inputs, is run only on MNIST and Fashion-MNIST.

\section{Full CIFAR-10 ablation}
\label{sec:supp-ablation}
\Cref{tab:ablation-full} is \cref{tab:ablation} with the rows that vary the width and fan-in of the second convolution.
A second convolution adds 4.38 points over one, and a third does not help (73.57\%) at 50.1k more LUTs; widening the second from 192 to 384 and 768 channels adds 1.74 and 1.30 points, and 1{,}536 channels add 0.37 for 27.0k more LUTs.
Fan-ins of $k_2=2$, 3 and 4 two-bit codes reach 72.40, 73.46 and 73.67\%; $k_2=3$ is the largest that fits one LUT (\cref{eq:fanin}), and $k_2=4$ needs 55.8k more LUTs.
Trading code width for fan-in, $k_2=6$ with 1-bit codes costs 0.77 points but saves 56.8k LUTs.

\begin{table}[t]
\centering
\footnotesize
\caption{CIFAR-10 ablations, one change each from the reference (conv2 768 channels, $k_2=3$, head $8000\to4000$ in one block), seed 42.
LUTs of the routed design; for the designs that do not route ($\times$), after placement.
Routes: $\checkmark$ placed and routed at 8.0\,ns without clock search (the penalty row at 5.0\,ns, the last row at 4.9\,ns); $\times$ does not route.
Blocked heads also use the position-major bus (\cref{sec:scale}); the last row is the CIFAR-10 design of \cref{tab:main}.}
\label{tab:ablation-full}
\setlength{\tabcolsep}{3pt}
\begin{tabular}{lrrr}
\toprule
Change & Acc.\ (\%) & LUTs & Routes \\
\midrule
Reference                                 & 73.46 & 194.6k & $\checkmark$ \\
\midrule
\multicolumn{4}{l}{\emph{Training}} \\
Hard top-$k$ cut (no annealing)           & 68.62 & 178.5k & $\checkmark$ \\
No S1 surrogate head                      & 72.06 & 189.4k & $\checkmark$ \\
Linear nodes                              & 73.04 & 189.2k & $\checkmark$ \\
Random sub-net init                       & 70.58 & 174.6k & $\checkmark$ \\
Random fixed taps                         & 71.38 & 214.5k & $\checkmark$ \\
Taps from one input channel               & 68.66 & 171.3k & $\checkmark$ \\
With rank-decay penalty~\cite{modi2026topk} & 72.63 & 186.8k & $\checkmark$ \\
\midrule
\multicolumn{4}{l}{\emph{Head}} \\
Random layer-1 wiring                     & 66.59 & 212.9k & $\times$ \\
Funnel LUT-tree head                      & 61.78 & 265.7k & $\checkmark$ \\
TT-Sparse head~\cite{soegeng2026ttsparse} & 45.40 & 64.8k  & $\checkmark$ \\
\midrule
\multicolumn{4}{l}{\emph{Architecture}} \\
Depth 1 (conv1 only)                      & 69.08 & 62.3k  & $\checkmark$ \\
Depth 3 (+ conv3, 768 ch)                 & 73.57 & 244.7k & $\checkmark$ \\
conv2 192 ch                              & 70.42 & 138.3k & $\checkmark$ \\
conv2 384 ch                              & 72.16 & 164.8k & $\checkmark$ \\
conv2 1{,}536 ch                          & 73.83 & 221.6k & $\checkmark$ \\
conv2 384 ch at $16\times16$              & 73.99 & 210.2k & $\checkmark$ \\
$k_2=2$ (4 bits)                          & 72.40 & 177.6k & $\checkmark$ \\
$k_2=4$ (8 bits)                          & 73.67 & 250.4k & $\checkmark$ \\
$k_2=6$ (1-bit codes)                     & 72.69 & 137.8k & $\checkmark$ \\
Head $4000\to2000$                        & 71.04 & 168.3k & $\checkmark$ \\
Head $16000\to8000$, 1 block              & 75.13 & 218.3k & $\times$ \\
Head $16000\to8000$, 2 blocks             & 74.68 & 211.1k & $\checkmark$ \\
Head $16000\to8000$, 4 blocks             & 73.93 & 210.6k & $\checkmark$ \\
Head $32000\to16000$, 1 block             & 76.68 & 250.8k & $\times$ \\
Head $32000\to16000$, 4 blocks            & 75.83 & 264.0k & $\checkmark$ \\
\bottomrule
\end{tabular}
\end{table}

\end{document}